\pdfoutput=1

\documentclass[11pt]{article}

\usepackage[preprint]{acl}

\usepackage{times}
\usepackage{latexsym}

\usepackage[T1]{fontenc}

\usepackage[utf8]{inputenc}

\usepackage{microtype}

\usepackage{inconsolata}

\usepackage{graphicx}

\usepackage{amsmath}
\usepackage{amssymb}
\usepackage{mathtools}
\usepackage{amsthm}

\usepackage{algorithm}
\usepackage{algorithmic}
\usepackage{hyperref}
\usepackage[capitalize,noabbrev]{cleveref}
\usepackage{booktabs}
\usepackage{subcaption}
\usepackage{xspace}

\newcommand{\argmax}{\operatornamewithlimits{argmax}}

\newcommand{\gemma}{\texttt{Gemma-2B}\xspace}
\newcommand{\gemmait}{\texttt{Gemma-2B-it}\xspace}
\newcommand{\pairrm}{\texttt{PairRM}\xspace}
\newcommand{\llama}{\texttt{Llama3-8B}\xspace}
\newcommand{\llamait}{\texttt{Llama3-8B-it}\xspace}

\begin{document}

\title{Semi-Supervised Reward Modeling via Iterative Self-Training}


\author{
 \textbf{Yifei He$^\ast$\textsuperscript{1}},
 \textbf{Haoxiang Wang$^\ast$\textsuperscript{1}},
 \textbf{Ziyan Jiang\textsuperscript{2}},
 \textbf{Alexandros Papangelis\textsuperscript{2}},
 \textbf{Han Zhao\textsuperscript{1,2}}
\\
 \textsuperscript{1}University of Illinois Urbana-Champaign
 \textsuperscript{2}Amazon \\
 \texttt{\{yifeihe3,hwang264,hanzhao\}@illinois.edu} \\
 \texttt{\{ziyjiang,papangea\}@amazon.com}
 }

\maketitle
\newcommand{\customfootnotetext}[2]{{%
		\renewcommand{\thefootnote}{#1}%
		\footnotetext[1]{#2}}}%
\customfootnotetext{$\ast$}{Equal contribution.}

\begin{abstract}
Reward models (RM) capture the values and preferences of humans and play a central role in Reinforcement Learning with Human Feedback (RLHF) to align pretrained large language models (LLMs). Traditionally, training these models relies on extensive human-annotated preference data, which poses significant challenges in terms of scalability and cost. To overcome these limitations, we propose Semi-Supervised Reward Modeling (SSRM), an approach that enhances RM training using unlabeled data. Given an unlabeled dataset, SSRM involves three key iterative steps: pseudo-labeling unlabeled examples, selecting high-confidence examples through a confidence threshold, and supervised finetuning on the refined dataset. Across extensive experiments on various model configurations, we demonstrate that SSRM significantly improves reward models without incurring additional labeling costs. Notably, SSRM can achieve performance comparable to models trained entirely on labeled data of equivalent volumes. Overall, SSRM substantially reduces the dependency on large volumes of human-annotated data, thereby decreasing the overall cost and time involved in training effective reward models.\footnote{Our code is available at \url{https://github.com/RLHFlow/RLHF-Reward-Modeling/tree/main/pair-pm}.}
\end{abstract}

\section{Introduction}

\begin{figure*}[t!]
    \centering
    \includegraphics[width=0.9\linewidth]{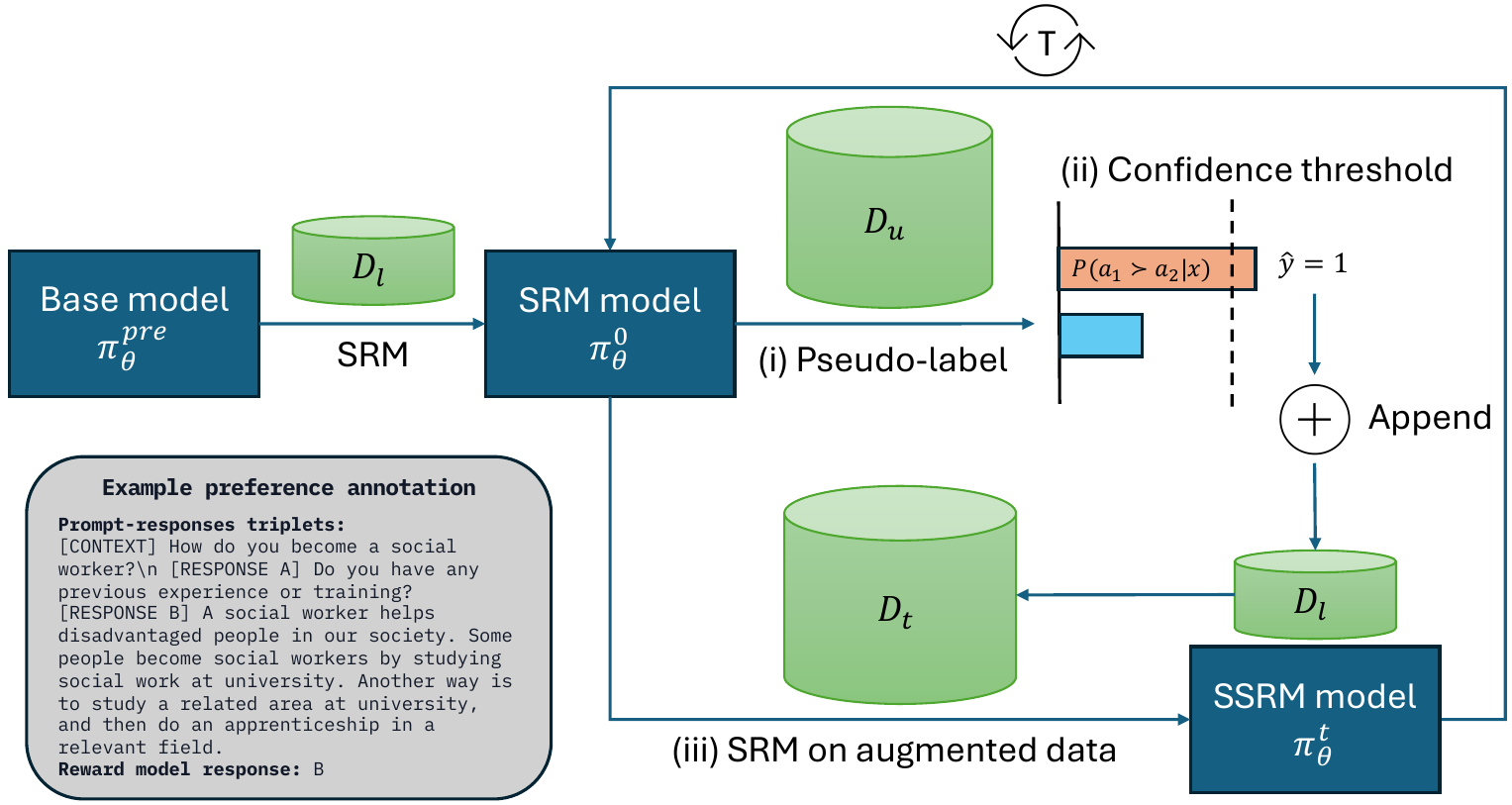}
    \caption{Semi-Supervised Reward Modeling (SSRM) enhances the ability of a language model to predict preferences using both labeled and unlabeled data. Given a pretrained model $\pi_\theta^\text{pre}$, a small labeled dataset $D_l$ and a large unlabeled dataset $D_u$, we first perform supervised reward modeling (SRM) on $D_l$ to obtain the SRM model $\pi_\theta^0$. Then, at each step $t$, we perform three steps: (i) Pseudo-labeling: assign pseudo-labels to examples in $D_u$. (ii) Confidence thresholding: given a prompt $x$ and two responses $a_1, a_2$, if the prediction confidence exceeds a preset threshold, append it to the labeled dataset to obtain $D_t$. (iii) SRM on augmented data: finetune the model on $D_t$.}
    \label{fig:ssrm}
\end{figure*}

Reinforcement Learning from Human Feedback (RLHF)~\citep{christiano2017deep,ziegler2019fine,ouyang2022training,bai2022constitutional} is central to the advancement of large language models (LLMs), including GPT-4~\citep{achiam2023gpt} and Claude~\citep{bai2022training}. In RLHF, a reward model is trained to capture common human values and preferences by learning from pairwise preference data. Subsequently, language models are refined based on the reward signal to ensure that their actions are aligned with human preferences. This enables language models to perform complex tasks with intricate and challenging objectives, including mathematical reasoning~\citep{wei2022chain,lewkowycz2022solving}, code generation~\citep{chen2021evaluating,li2022competition}, summarization~\citep{ouyang2022training}, among others.

While the resulting aligned language models produced by RLHF receive considerable attention, the importance of reward models is often overlooked. The accuracy of reward models in capturing human preferences is crucial for the effectiveness of RLHF. Moreover, beyond the application in RLHF, the ability to annotate preferences for prompt-response data enables reward models to serve as a valuable tool for a broader range of alignment approaches, including Rejection Sampling Finetuning (RSF)~\citep{dong2023raft,gulcehre2023reinforced,yuan2024self} and Direct Preference Optimization (DPO)~\citep{rafailov2024direct}. \looseness=-1

However, training reward models necessitates a significant volume of human-annotated preference data. Furthermore, in real-world applications, it is typical to encounter examples that deviate from the predefined training domains, and it is impractical to gather annotated preference data for every conceivable domain, presenting a significant barrier to the deployment of reward models. This challenge underscores the need to explore methods for enhancing reward models without relying extensively on large datasets of human-annotated preferences. 

To enhance the efficiency of data utilization in reward model training, we propose Semi-Supervised Reward Modeling (SSRM), which efficiently utilizes unlabelled data to improve reward models. Our methodology stems from seminal works in the field of boosting~\citep{schapire1990strength,freund1995boosting} and semi-supervised learning~\citep{seeger2000learning,grandvalet2004semi}, which aims at converting weak models to strong models with minimal requirement of labeled data. Given a pretrained language model and an unlabeled dataset with prompt-response pairs, SSRM iteratively executes the following steps: (i) Pseudo-labeling: assign pseudo-labels to the unlabeled examples based on their predicted preferences, (ii) Confidence thresholding: employ confidence threshold to selectively retain examples where the model exhibits high certainty in its predictions, (iii) Supervised reward-modeling (SRM): finetune on the filtered subset of data to enhance the reward model. 

Through extensive experiments on various language models with parameter counts ranging from 0.4B to 8B, we demonstrate that SSRM effectively improves the performance of reward models without additional labeling costs. Notably, our findings reveal that reward models trained using SSRM exhibit performance closely approaching models trained entirely through traditional supervised methods on equivalent volumes of data. This underscores the efficacy of SSRM in utilizing unlabeled data to mirror the learning gains typically achieved only with labeled datasets.

\section{Semi-Supervised Reward Modeling}

Semi-supervised reward modeling (SSRM) utilizes a small amount of labeled data and a large amount of unlabeled data to efficiently enhance the capabilities of reward models. In \Cref{sec:pref_model}, we introduce reward modeling and our training recipe. In \Cref{sec:st}, we detail the self-training procedure in the context of reward modeling.

\begin{algorithm}[t!]
\caption{Semi-Supervised Reward Modeling}\label{algo:main}
    \begin{algorithmic}[1]
        \STATE \textbf{Input: }a pretrained model $\pi_\theta^{\text{pre}}$, a labeled dataset $D_l=\{(x^{(i)},a_1^{(i)},a_2^{(i)}, y^{(i)})\}_{i=1}^m$, an unlabeled dataset $D_u=\{(x^{(i)},a_1^{(i)},a_2^{(i)})\}_{i=1}^n$, confidence threshold $s$, iteration number $T$
        \STATE \textbf{Output: }improved reward model $\pi_\theta^T$
        \STATE \texttt{// Initial SRM}
        \STATE $\pi_\theta^0 \leftarrow \text{Update}(\pi_\theta^{\text{pre}}, \ell_\text{SRM}, D_l)$
        \FOR{$t=1,2,\cdots,T-1$}
            \STATE Initialize $D_t = D_l$
            \FOR{$i=1,2\cdots,n$}
                \STATE \texttt{// Pseudo label}
                \STATE $\hat{y}^{(i)}=\argmax_y \pi_\theta^t\left(y|x^{(i)},a_1^{(i)},a_2^{(i)}\right)$
                \STATE \texttt{// Confidence threshold}
                \IF{$\pi_\theta^t \left(\hat{y}^{(i)}|\mathbb{T}\left(x^{(i)},a_1^{(i)},a_2^{(i)}\right)\right) \geq s$}
                    \STATE $D_t \leftarrow D_t \cup \{(x^{(i)},a_1^{(i)},a_2^{(i)},\hat{y}^{(i)})\}$
                \ENDIF
            \ENDFOR
            \STATE $\pi_\theta^{t+1} \leftarrow \text{Update}(\pi_\theta^t, \ell_\text{SRM}, D_t)$
        \ENDFOR
    \end{algorithmic}
\end{algorithm}

\subsection{Reward Model} \label{sec:pref_model}
Reward modeling aims to encode human values by predicting the preference of a pair of responses given the same prompt. For reward models, two prominent types are widely recognized: Bradley-Terry models~\citep{bradley1952rank} and preference models. Preference models have the flexibility to provide a more generalizable approach for capturing preferences as compared to Bradley-Terry models~\citep{munos2023nash}. Furthermore, they can be trained in a more computationally efficient manner. Consequently, we focus on preference models and employ them in our framework. 

A preference model takes a prompt $x$ and two responses $a_1,a_2$ as inputs and predicts the preference score $P(a_1 \succ a_2|x)$, indicating the preference of response $a_1$ over response $a_2$ given $x$. In the implementation, we adopt the methodology described by \citet{zhao2023slic}, which casts preference modeling as an instruction following task by leveraging the capabilities of LLMs for next-token prediction. Each preference pair takes the form $(x,a_1,a_2,y)$, where $y\in\{A,B\}$ denotes whether the first or the second response is more preferable. The instruction template $\mathbb{T}(x,a_1,a_2)$ is formatted as \texttt{[CONTEXT]\{x\}[RESPONSE A]\{$a_1$\}[RESPONSE B]\{$a_2$\}}, and the target is the index for the preferred response (example shown in \Cref{fig:ssrm}). To mitigate the positional bias, i.e., the tendency for the ordering of responses to influence preference, we randomize their order during data preparation. To differentiate from supervised finetuning (SFT), which refers to finetuning for general instruction following, we term supervised finetuning on preference data as supervised reward modeling (SRM). At the end, we use SRM to train the reward model
\begin{align*}
    \ell_{\text{SRM}}(\pi_\theta)=-\mathbb{E}_{(x,a_1,a_2,y)}[\log \pi_\theta(y|\mathbb{T}(x,a_1,a_2))].
\end{align*}
During inference, we directly use the probability of decoding the correct label, i.e., $\pi_\theta(y|\mathbb{T}(x,a_1,a_2))$, as the preference score.

\subsection{Iterative Self-Training} \label{sec:st}
Training reward models requires pairwise data annotated with preference, consuming significant human efforts and resources. On the other hand, unlabeled data is easily accessible as language models can generate diverse responses given prompts. Therefore, to reduce the labeling cost for preference learning, we propose to utilize self-training~\citep{grandvalet2004semi,lee2013pseudo}. Specifically, we leverage confident predictions of a model to produce pseudo-labels for the unlabeled data and train on this augmented dataset iteratively. In the context of reward modeling, a labeled dataset takes the form $D_l=\{(x^{(i)},a_1^{(i)},a_2^{(i)},y^{(i)})\}_{i=1}^m$, and an unlabeled dataset takes the form $D_u=\{(x^{(i)},a_1^{(i)},a_2^{(i)})\}_{i=1}^n$. Typically, the volume of unlabeled data vastly exceeds that of labeled data ($n\gg m$), providing a rich resource for augmenting the training dataset through self-training. The detailed steps of applying self-training in reward modeling are as follows.


\paragraph{Supervised training} Initially, we train a reward model on the labeled dataset $D_l$ using SRM mentioned in the previous section
\begin{align*}
    \pi_\theta^0=\argmax_\theta \sum_{i=1}^m \log \pi_\theta\left(y^{(i)}|\mathbb{T}\left(x^{(i)},a_1^{(i)},a_2^{(i)}\right)\right).
\end{align*}
This supervised model serves as a starting point of the self-training pipeline. The subsequent steps update upon this supervised model iteratively.

\paragraph{Pseudo-labeling} At each iteration $t$, we assign pseudo-labels to unlabeled data in $D_u$ based on the model predictions. For each data point in $D_u$, we select the response with the higher preference score as the pseudo-label
\begin{align*}
    \hat{y}^{(i)}=\argmax_y \pi_\theta^t\left(y|x^{(i)},a_1^{(i)},a_2^{(i)}\right).
\end{align*}
Here, we employ hard labeling: we pseudo-label data points in a binary way, instead of a probabilistic label based on the output logits.

\paragraph{Confidence thresholding} After pseudo-labeling, it is crucial not to directly use the entirety of pseudo-labeled data for self-training, as doing so will result in a final model with identical performance as the initial model~\citep{chapelle2006semi}. Thus, we only select those data where the model exhibits high confidence. In the context of reward modeling, we compute the confidence based on the preference score of the assigned pseudo-label
\begin{align*}
    \max_y \pi_\theta^t \left(y|\mathbb{T}\left(x^{(i)},a_1^{(i)},a_2^{(i)}\right)\right).
\end{align*}
For a preset confidence threshold $s$, we only retain the pseudo-labeled data with confidence above the threshold, which are then combined with the labeled data to form the new training set for the current iteration, denoted as $D_t$.

\paragraph{Model update} Following the dataset construction, we perform another round of SRM on $D_t$ to get the updated model.

These steps are repeated for a preset number of iterations, with the entire procedure detailed in \Cref{algo:main}. This approach efficiently leverages unlabeled data, reducing reliance on expensive labeled datasets and iteratively enhancing the model’s performance in predicting human preferences. \looseness=-1

\section{Experiments}
Our experiments are designed to evaluate the scalability and efficiency of SSRM across a spectrum of model sizes and configurations. We use a confidence threshold of 0.8, and more implementation details can be found in \Cref{appendix:exp}.

\subsection{Setup}

\paragraph{Models}~We utilize three models to ensure a comprehensive assessment: \pairrm~\citep{jiang2023llm}, \gemmait~\citep{team2024gemma} and \texttt{Llama3-8b-it}~\citep{llama3}, with 0.4B, 2B and 8B parameters respectively. Note that \pairrm is an encoder-based model specifically designed for reward modeling, so we follow the training methodology in \citet{jiang2023llm} instead of training it through the mentioned SRM approach in \Cref{sec:pref_model}, which is only applicable for language models with generation capabilities. 

\paragraph{Datasets}~To fit the specific formatting requirements of the models, we use two datasets for training. For experiments involving \gemma and \llama, we follow \citet{dong2024rlhf} to use a mixture of 8 open-source datasets for reward model training: HH-RLHF~\citep{bai2022training}, SHP~\cite{ethayarajh2022understanding}, HelpSteer~\citep{wang2023helpsteer}, PKU-SafeRLHF~\citep{ji2024beavertails}, UltraFeedback~\citep{cui2023ultrafeedback}, UltraInteract~\citep{yuan2024advancing}, Distilabel-Capybara~\citep{daniele2023amplify-instruct} and Distilabel-Orca~\citep{lian2023openorca}. Collectively, these datasets provide a rich and diverse corpus of 700K prompt-response triplets. Initially, these datasets come with ground-truth labels, but in our semi-supervised learning setup, we utilize only a tiny portion of the labels for the initial stages of supervised training (SRM) as described in \Cref{sec:st}. The remainder of the data, which constitutes the majority, is used unlabeled.\looseness=-1

For experiments involving \pairrm, we use the OpenHermesPreferences dataset~\citep{open_hermes_preferences}, which contains 990k prompt-response triplets. This dataset is selected due to its compatibility with the specific data formatting requirements of \texttt{PairRM}, ensuring optimal training and performance evaluation conditions. We use the dataset in a purely unsupervised manner.

\begin{table}[t!]
    \centering
    \scalebox{0.8}{
    \begin{tabular}{cc}
        \toprule
        Model & \# Data for SRM \\
        \midrule
        \gemmait & 175K \\
        \llamait & 43.75K \\
        \pairrm & 0 \\
        \bottomrule
    \end{tabular}}
    \caption{Number of labeled data used for SRM.}\label{tab:n_data}
\end{table}

\paragraph{Data Splitting}~We detail our approach to partitioning the datasets into labeled and unlabeled segments for training (summarized in \Cref{tab:n_data}). Both \gemmait and \llamait are general-purpose language models that have not been tuned on preference datasets, so we first perform supervised training on a small labeled subset of the preference dataset. Specifically, we train \gemmait on one-fourth of the dataset, and train \llamait on one-sixteenth of the dataset. This distinction in the volume of data used for initial training reflects their inherent differences in model capabilities and processing capacity. For instance, training \llamait with a larger share of the dataset (such as one-fourth) tends to lead to an oversaturation of its learning capacity, thereby leaving minimal room for potential gains from subsequent exposure to unlabeled data (more detailed discussion in \Cref{appendix:exp}). This SRM process with limited labeled data aims to equip the models with a basic understanding of preference learning, ensuring that the pseudo-labeling conducted in the initial iteration of SSRM has reasonable accuracy. 

Conversely, \pairrm operates differently. As an encoder-based model explicitly developed for reward modeling, \pairrm is already equipped with advanced capabilities for preference prediction. Therefore, we exempt it from additional supervised training on preference data, skipping the SRM step described in \Cref{sec:pref_model}. Instead, \pairrm is employed directly in our SSRM setup for pseudo-labeling, leveraging its innate abilities to process and evaluate preference data effectively.

\begin{table*}[t!]
    \centering
    \scalebox{0.8}{
    \begin{tabular}{lcccccc}
    \toprule
                     & Chat   & Chat Hard & Safety & Reasoning & Average  & \# Data (Pseudo-labeled portion) \\
    \midrule
    \gemmait      & 52.51 & \textbf{50.00}    & 42.36 & 48.50    & 48.34 & 0 (0)                \\
    Partial SRM  &  90.78 &  35.96    &  31.38 &  51.61    & 52.44 & 175K (0)                \\
    \midrule
    SSRM [t=1]     &  \textbf{95.25} &  37.06    &  47.70 &  41.37    &  55.34 & 310K ($43.5\%$)             \\
    SSRM [t=2]     &  94.41 &  37.06    &  49.39 &  \textbf{68.10}    &  62.24 & 406.3K ($56.9\%$)           \\
    SSRM [t=3]     &  94.41 &  35.65    &  \textbf{53.79} &  66.33    & \textbf{ 62.54} & 402.7K ($56.5\%$)          \\
    \midrule
    Full SRM &  94.97 &  37.50    &  61.76 &  68.81    &  65.76 & 770K (0)               \\
    \bottomrule
    \end{tabular}}
    \caption{RewardBench evaluation for \gemma models. We start with SRM on one-fourth of the dataset. The overall performance substantially improves through SSRM, where the drop in Chat Hard results from its conflict with Chat. Notably, the SSRM performance approaches that of the model trained in a purely supervised manner.}\label{tab:gemma}
\end{table*}

\paragraph{Evaluation}~The efficacy of our reward models is assessed using RewardBench~\citep{lambert2024rewardbench}, a comprehensive evaluation benchmark for reward modeling. This benchmark consists of 2,985 prompt-chosen-rejected triplets, which encompass a range of critical evaluation criteria such as instruction following, reasoning, and safety. Specifically, it incorporates common LLM evaluation benchmarks formatted for reward model assessment, including MT-Bench~\citep{zheng2024judging}, AlpacaEval~\citep{li2023alpacaeval} and HumanEval~\citep{chen2021evaluating}. The primary metric of evaluation is the accuracy of predicting the chosen response. Performance on RewardBench serves as a strong and direct indicator of the reward model's capability to align policy language models effectively.

\subsection{Benchmark Evaluation}
\paragraph{Gemma-2B} We report the RewardBench evaluation on \gemma in \Cref{tab:gemma}. Initially, the out-of-the-box performance of \gemmait, detailed in the first row of the table, serves as a baseline that reveals the model's rudimentary capabilities as a reward model. At this stage, its performance across all four categories closely resembles random guessing, indicating significant room for improvement. The second row represents the model trained with only a quarter of the labeled data, which shows marginal improvements. This underscores the challenges in significantly enhancing the performance of preference learning with limited labeled data. 

As detailed in the subsequent rows, we apply the SSRM process described in \Cref{algo:main} for three iterations. Throughout these iterations, we observe a consistent improvement in performance, with the most significant gains occurring between the first and second iterations. This notable enhancement highlights the benefits of the iterative self-training approach. Note that the performance plateaus after the second iteration, as indicated by the similar metrics in the $t=2$ and $t=3$ rows. This suggests that once the learning capacity of the model is saturated, additional iterations may not yield further substantial gains. \looseness=-1

We note a decrease in performance within the Chat Hard category, which can be attributed to the inherent biases in different chat categories. As observed in \citet{dong2024rlhf,wang2024helpsteer2}, the Chat category is verbosity biased and the Chat Hard category is simplicity biased, each favoring responses of different lengths. These biases establish a competitive dynamic where improvements in one category can inadvertently lead to declines in the other. This relationship accounts for the observed drop in Chat Hard performance, but we see significant gains in the Chat category. The conflict of these biases is particularly pronounced in smaller models, where the limited model capacity must balance the conflicting tasks, exacerbating trade-offs. Conversely, in larger models like \llama, this conflict tends to be less obvious.

The final row reports the oracle performance of the full SRM \gemmait on the complete dataset with ground-truth labels. The model with SSRM achieves performance metrics closely approaching those of the full SRM model, despite using only one-fourth of the labeled data. This efficiency is significant, highlighting SSRM's effectiveness in leveraging pseudo-labels to enhance model performance without extensive reliance on labeled data.

We also report the number of training data used in each SSRM iteration in the last column. The proportion of pseudo-labels shows an increasing trend, illustrating the model's growing confidence in its predictions, which allows it to self-train on an expanding pool of data. A more detailed discussion on prediction confidence is presented in \Cref{sec:analysis}. We report the number of data used for subsequent models in \Cref{appendix:exp} due to space limit.

\begin{table}[t!]
    \centering
    \scalebox{0.7}{
    \begin{tabular}{lccccc}
    \toprule
                  & Chat   & Chat hard & Safety & Reasoning & Average \\
    \midrule
    \llamait        &  44.69 &  53.29    &  59.15 &  50.82    &  51.99  \\
    Partial SRM        &  96.09 &  40.79    &  62.37 &  76.11    &  68.83  \\
    \midrule
    SSRM [t=1]       &  97.77 &  52.85    &  74.04 &  91.25    &  78.98  \\
    SSRM [t=2]       &  96.65 &  \textbf{64.04} &  83.42 &  87.03 &  82.78  \\
    SSRM [t=3]   &  \textbf{98.32} &  59.21 &  \textbf{85.68} &  \textbf{93.57} & \textbf{ 84.19}\\
    \midrule
    Full SRM &  98.60 &  65.35 &  88.81 &  92.07 &  86.21 \\
    \bottomrule
    \end{tabular}}
    \caption{RewardBench evaluation for \llama models. We start with SRM on one-sixteenth of the dataset. SSRM significantly improves the performance in all categories. Different from \gemma, both Chat and Chat hard improve as a result of larger model capacity.}\label{tab:llama}
\end{table}

\paragraph{Llama3-8B}~We report the results on \llama in \Cref{tab:llama}. With only one-sixteenth of the full dataset, the partial SRM performance already gets a noticeable boost. This substantial early gain, compared to the marginal improvements seen with \texttt{Gemma-2B} under similar conditions, underscores the greater potential of models with larger parameter counts to leverage limited data effectively. 

After applying SSRM, the performance of the model continues to improve. The improvement is particularly significant at the first iteration of SSRM, which indicates substantial gains from incorporating the self-training approach. As SSRM progresses through additional iterations, we observe further performance improvements, with notable advancements in categories like Safety and Reasoning, highlighting the model's improved reliability in sensitive scenarios. Similar to the observation in \gemma experiments, the performance with SSRM approaches that of the full SRM model, with only one-sixteenth of the labeled data. 

\begin{table}[t!]
    \centering
    \scalebox{0.7}{
    \begin{tabular}{lccccc}
    \toprule
                  & Chat   & Chat hard & Safety & Reasoning & Average \\
    \midrule
    \pairrm        &  90.22 &  \textbf{53.29}    &  39.80 &  48.80    &  58.03  \\
    \midrule
    SSRM [t=1]       &  \textbf{93.02} &  43.20    &  84.10 &  54.44    &  68.69  \\
    SSRM [t=2]       &  92.45 &  38.38    &  86.51 &  \textbf{58.56}    &  68.98  \\
    SSRM [t=3] &  91.51 &  42.98    &  \textbf{88.91} &  57.09    & \textbf{ 70.12}  \\
    \bottomrule
    \end{tabular}}
    \caption{RewardBench evaluation for \pairrm models. No SRM step is performed as \pairrm is a reward model. SSRM consistently enhances the model, showing its effectiveness for small encoder-based models.\looseness=-1}\label{tab:pairrm}
\end{table}

\paragraph{PairRM}~We report the results on \pairrm in \Cref{tab:pairrm}. Initially, since \pairrm is specifically trained for preference learning, it demonstrates a strong overall capability as a reward model. The performance is noticeably better compared with \gemmait and \llamait, despite these models possessing significantly more parameters. This difference emphasizes the distinct requirements and capabilities needed for preference learning as opposed to general language modeling tasks. However, its performance in Safety and Reasoning is considerably lower, indicating limitations in handling nuanced content out of the box. 

SSRM in subsequent rows reveals a clear trend of performance enhancement across iterations. The performance improvement is especially noticeable in the Safety category, where the original \pairrm underperforms. This improvement likely indicates an initial deficiency in safety-related data during the original \pairrm training, a gap which our semi-supervised approach begins to fill by leveraging unsupervised data. Like the observations with \llama, the most significant gains are observed at the first iteration of SSRM, highlighting the immediate impact of the semi-supervised learning process. Similar to our previous finding, the performance gain plateaus with more iterations.

Overall, the empirical results across models of varying sizes confirm the versatility and efficiency of SSRM in improving reward model performance. By effectively employing unlabeled data, SSRM not only enhances model capabilities but also presents a cost-effective training alternative to traditional fully supervised methods. These advantages are especially important in scenarios where acquiring extensive labeled datasets is impractical due to resource constraints.

\begin{table}[t!]
    \centering
    \scalebox{0.7}{
    \begin{tabular}{lc}
    \toprule
    Policy model & MT-bench score \\
    \midrule
    Unaligned \gemma & 1.41 \\
    DPO aligned by \gemmait [Partial SRM] & 1.76 \\
    DPO aligned by \gemmait [SSRM] & \textbf{2.29} \\
    \bottomrule
    \end{tabular}}
    \caption{MT-Bench evaluation for the pretrained \gemma aligned by different reward models. SSRM significantly enhances the model.}\label{tab:policy_lm}
\end{table}

\subsection{Evaluation of the aligned LM}
We demonstrate the effectiveness of SSRM-enhanced reward models in aligning LMs. Our experiment compares two models: SSRM-Gemma-2b-it (SSRM [t=3] in \Cref{tab:gemma}) and partial SRM-Gemma-2b-it (partial SRM in \Cref{tab:gemma}). Both models are used to pseudo-label the same set of data described in Section 3.1. We then perform one iteration of DPO on each of the resulting pseudo-labeled datasets to align a \gemma model (pretrained, not instruction-tuned). We report the GPT-4 judgment scores of the aligned model on MT-bench in \Cref{tab:policy_lm}. The policy model aligned by the SSRM model significantly outperforms its partial SRM counterpart, demonstrating that the enhanced reward modeling abilities directly contribute to more effective alignment of the policy model. These findings align with our RewardBench evaluation, further confirming that stronger reward modeling capabilities translate into improved performance in the aligned policy model.

\subsection{Empirical Analysis} \label{sec:analysis}

\begin{figure}[t!]
    \centering
    \begin{subfigure}[t]{0.23\textwidth}
        \centering
        \includegraphics[width=\linewidth]{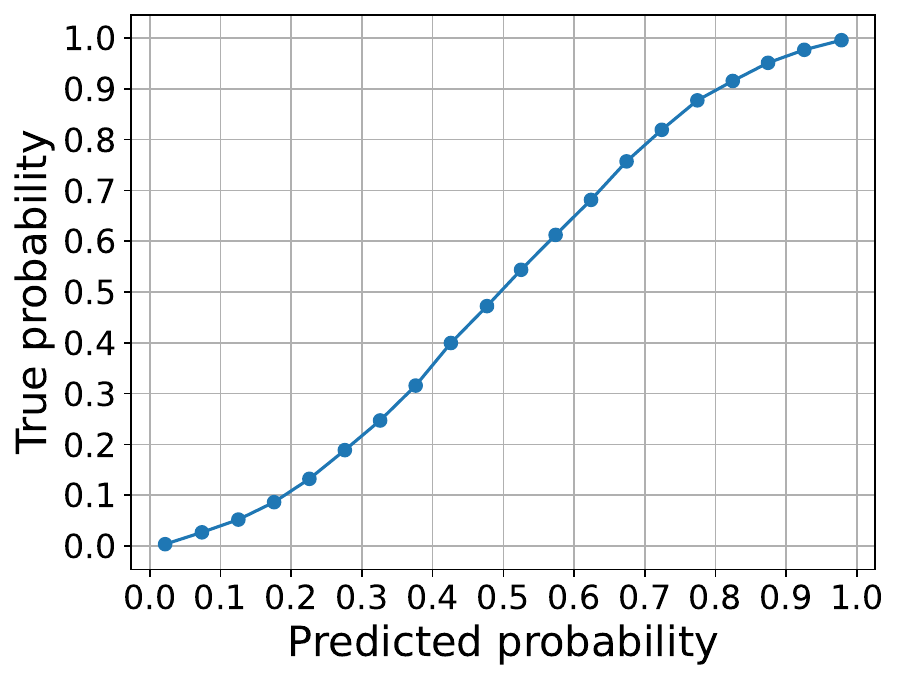}
        \caption{Partial SRM.}
        \label{fig:cali-sft}
    \end{subfigure}%
    ~
    \begin{subfigure}[t]{0.23\textwidth}
        \centering
        \includegraphics[width=\linewidth]{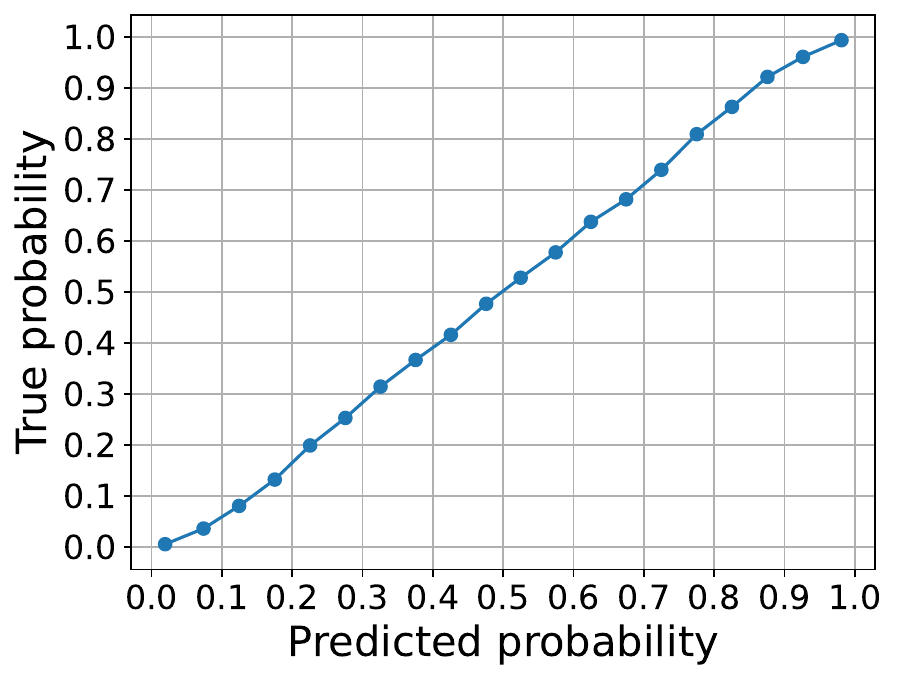}
        \caption{\small{SSRM [t=3].}}
        \label{fig:cali-iter3}
    \end{subfigure}
    \caption{The \gemma undergone three iterations of SSRM demonstrates better calibration, especially in the high-confidence score range, showing the effectiveness of confidence thresholding.}
    \label{fig:cali}
\end{figure}

\paragraph{Calibration}~We show how SSRM improves the calibration of \gemma models by comparing the calibration curves before and after SSRM in \Cref{fig:cali}. Calibration curves, which plot the true probability against the predicted probability, serve as indicators of how well the predicted probabilities of a model represent the actual outcomes. A perfectly calibrated model would have all points lying on the diagonal line. For the model after partial SRM (\Cref{fig:cali-sft}), the curve deviates from the diagonal, especially in the mid-range probabilities between 0.2 and 0.8. This deviation suggests that the model tends to be under-confident, as it predicts lower probabilities than the actual likelihood of the correct outcomes. In comparison, the model after three iterations of SSRM (\Cref{fig:cali-iter3}) shows a curve that adheres more closely to the diagonal across the entire probability spectrum. This closer alignment indicates that the SSRM \gemma model's predictions are more reliable and accurately reflect the true likelihoods of outcomes. The improvement implies that the SSRM has effectively used unlabeled data to correct the under-confidence and improve the overall prediction accuracy. Moreover, the notable improvement in calibration at higher confidence scores underscores the effectiveness of confidence thresholding based on the predicted probability. \looseness=-1

\begin{figure}[t!]
    \centering
    \begin{subfigure}[t]{0.23\textwidth}
        \centering
        \includegraphics[width=\linewidth]{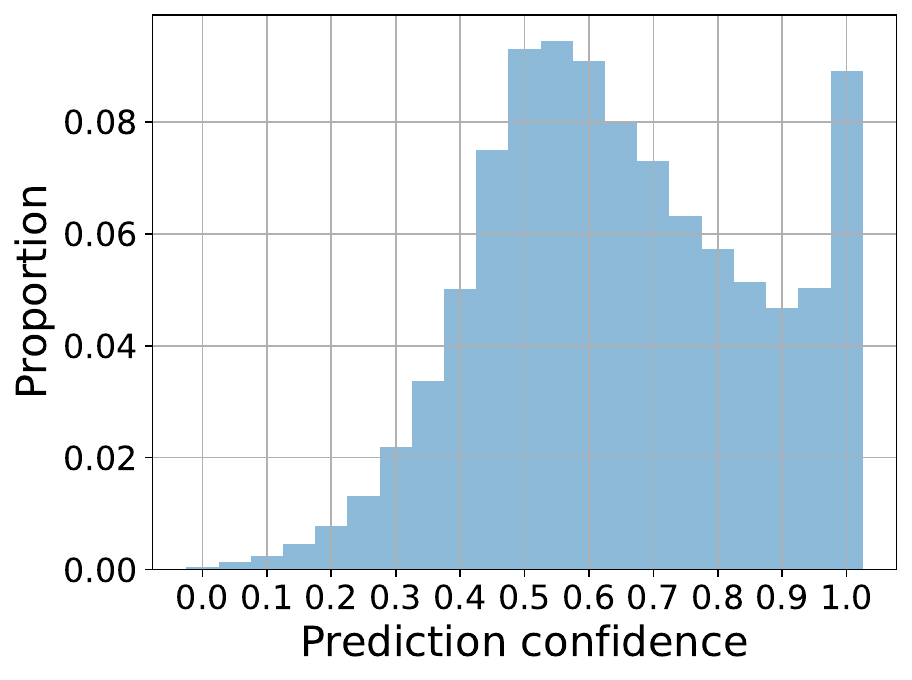}
        \caption{Partial SRM.}
    \end{subfigure}%
    ~
    \begin{subfigure}[t]{0.23\textwidth}
        \centering
        \includegraphics[width=\linewidth]{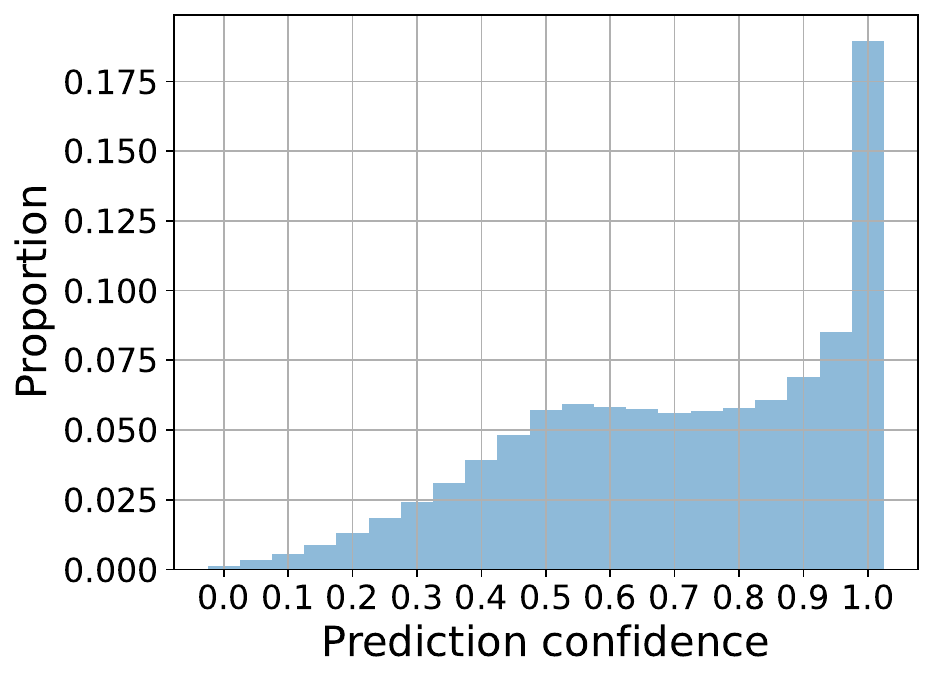}
        \caption{SSRM [t=3].}
        \label{fig:conf-iter3}
    \end{subfigure}
    \caption{The prediction confidence noticeably improves on \gemma models after three iterations of SSRM. Combined with better calibration, it shows the prediction more accurately reflects the actual outcome.}
    \label{fig:conf}
\end{figure}

\paragraph{Prediction confidence}~We show how SSRM improves the prediction confidence of the \gemma models in \Cref{fig:conf}. The prediction confidence is computed as the predicted probability of the ground-truth label, i.e., $\pi_\theta(y|\mathbb{T}(x,a_1,a_2))$. Initially, after SRM on a quarter of the dataset, the prediction confidence on remaining samples are relatively low, with the majority concentrating around 0.5. This suggests that the model's responses on many samples are akin to random guesses at this stage. However, subsequent application of SSRM after three iterations markedly shifts the confidence distribution towards the right, as depicted in \Cref{fig:conf-iter3}. This shift indicates a significant increase in the model’s prediction confidence at the dataset level. Importantly, this confidence increase is not merely a case of the model becoming more confidently incorrect. Rather, it is backed by improved model calibration, as previously analyzed. The enhanced model calibration indicates the correctness of confidence scores, hence higher confidence scores indeed imply more accurate pseudo-labels.


\begin{figure}[t!]
    \centering
    \includegraphics[width=0.8\linewidth]{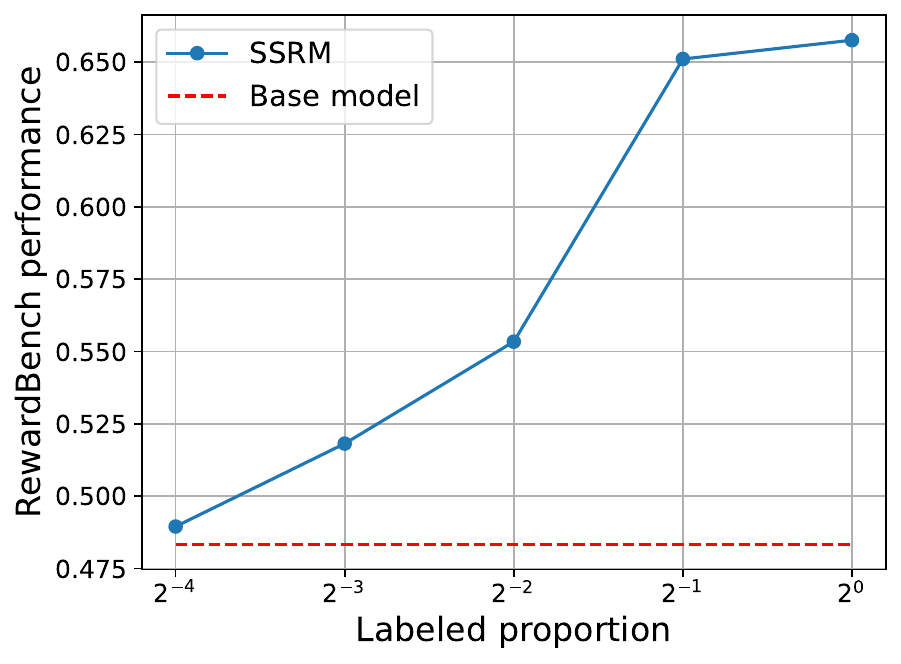}
    \caption{With more labeled data, the performance of SSRM consistently increases.}
    \label{fig:n_data}
\end{figure}

\paragraph{Number of labeled data}~In \Cref{fig:n_data}, we plot the effect of varying proportions of labeled data on the performance of SSRM applied to the \gemma model. In this experiment, SSRM is conducted over a single iteration. The x-axis represents different fractions of the dataset that are labeled, ranging from one-sixteenth to fully labeled. The horizontal dashed line indicates the baseline performance of \gemmait out of the box. The performance of the SSRM-enhanced model improves consistently as the amount of labeled data increases. Notably, when half of the dataset is labeled, the SSRM model's performance nearly matches that achieved by finetuning on a fully supervised dataset. This trend underscores the data efficiency of SSRM, demonstrating its ability to leverage increasing amounts of labeled data effectively.

\section{Related Works}
\paragraph{Semi-Supervised Learning}~Semi-supervised learning (SSL) aims at blending both labeled and unlabeled data to improve the performance of a model. SSL has been widely applied to NLP tasks, including text classification~\citep{gururangan-etal-2019-variational,meng2020text}, dependency parsing~\citep{li-etal-2019-semi-supervised-domain}, NER~\citep{chen-etal-2020-local}, and sequence generation~\citep{He2020Revisiting}. A cornerstone of SSL is self-training (ST), also known as pseudo-labeling~\citep{grandvalet2004semi,lee2013pseudo}. During ST, models assign pseudo-labels to unlabeled data and subsequently train on this augmented dataset in an iterative manner. This method has inspired numerous adaptations, including mean teacher~\citep{tarvainen2017mean}, noisy student~\citep{xie2020self} and FixMatch~\citep{sohn2020fixmatch}. These ST-based techniques have proven particularly effective in applications where labeled data is scarce or expensive to obtain, such as unsupervised domain adaptation~\citep{li-etal-2019-semi-supervised-domain,kumar2020understanding,he2023gradual}. These methods have also been applied in preference learning~\citep{cao2021weak,park2021surf} in the general RL settings, but mainly with applications in the vision and robotics domains. In this work, we extend these methodologies to the domain of reward modeling in RLHF, aiming to address the specific challenges posed by the high dependency on labeled data in training language models.

\paragraph{Reinforcement Learning from Human Feedback (RLHF)}~Traditional RLHF approaches~\citep{christiano2017deep,ziegler2019fine,ouyang2022training,bai2022constitutional} involve training a reward model using human-annotated preference data, and then use the reward signal provided by the reward model to align the behavior of language models with human values, using RL techniques such as Proximal Policy Optimization (PPO)~\citep{schulman2017proximal}. However, the training of PPO is challenging due to its inefficiency and instability compared with SFT. This drawback makes the results of PPO-based RLHF model such as ChatGPT~\citep{achiam2023gpt} largely non-reproduced in the open-source community. To overcome these limitations, alternative approaches like Rejection Sampling Finetuning (RSF)~\citep{dong2023raft,gulcehre2023reinforced,yuan2024self} have emerged. RSF learns from high-quality examples chosen by a reward model, then SFT on them. The pipeline has also shown success in aligning language models, including Llama-2~\citep{touvron2023llama}. Both RSF and RLHF operate in an online manner, meaning that the preference is judged based on the generation of the models. Alternatively, Direct Preference Optimization (DPO)~\citep{rafailov2024direct} aligns models in an offline manner, simplifying the process by tuning models directly on a curated preference dataset without the need for a separate reward model. \looseness=-1

Our proposed SSRM is beneficial across both online and offline frameworks. In the online scenarios, SSRM produces an enhanced reward model without requirement on more labeled data. In the offline case, the improved reward model can serve as an annotator that produces high-quality preference dataset for the subsequent procedure such as DPO. This versatility makes SSRM a valuable asset in advancing the field of RLHF.

\paragraph{Reinforcement Learning from Artificial Intelligence Feedback (RLAIF)}~The training of reward models in the traditional RLHF process requires substantial human-annotated preference data, incurring high cost for labeling. To mitigate these challenges, researchers have explored using language models themselves as a source of preference feedback, thus replacing the need for extensive human intervention. For instance, \citet{bai2022training} uses a language model to provide feedback and refine responses, enhancing reward models. This approach has been further validated by \citet{lee2023rlaif}, which demonstrates that AI-generated feedback can achieve comparable results to human feedback while significantly reducing the need for human labor. More recent advancements involve using LLMs directly as judges in a process known as LLM-as-a-Judge prompting \citep{li2023alpacaeval,dubois2024alpacafarm,bai2024benchmarking}, where LLMs are provided with specific rubrics and tasked with scoring the responses accordingly. 

While RLAIF has shown potential by leveraging powerful LLMs to simulate human feedback, this approach often relies on highly capable and therefore expensive LLMs (e.g., GPT-4) to achieve human-comparable feedback quality. This dependency means that querying these advanced models still incurs significant costs, limiting the accessibility and scalability of RLAIF approaches. In contrast, SSRM is designed to enhance language models of various sizes and capacities, including models with size as small as 0.4B parameters. This method allows even smaller, less resource-intensive models to improve their performance significantly, thus broadening the accessibility of effective training techniques and substantially reducing the costs associated with acquiring accurate feedback.

\section{Conclusion}
In this work, we address the substantial reliance of reward model training on extensive human-annotated preference data. We introduce Semi-Supervised Reward Modeling (SSRM), a method that mitigates this issue by effectively utilizing unlabeled data in conjunction with limited labeled data. The SSRM process consists of three primary steps: pseudo-labeling the unlabeled examples, selecting those examples with high prediction confidence, and finetuning the model on this augmented dataset. Our extensive experimental evaluations across models of varying sizes demonstrate that SSRM significantly enhances the performance of reward models with minimal requirement of labeled data. Furthermore, the performance of models trained using SSRM closely approaches that of models trained with equivalent volumes of fully supervised data. This performance is further supported by detailed analyses of calibration and prediction confidence, which underscore the robustness and effectiveness of SSRM. Overall, SSRM offers a highly efficient strategy for improving reward models, significantly reducing the need for costly and time-consuming data annotation. 

\section*{Limitations}
One constraint of the SSRM framework is its dependency on the initial reward modeling capabilities of the model. Especially when labeled data is scarce, the model after only the initial stage of supervised training might not acquire enough knowledge to accurately assign pseudo-labels to the unlabeled dataset. This limitation poses a risk of generating low-quality pseudo-labels, which could potentially propagate errors through the training process. Nevertheless, it is important to note that we incorporate a confidence thresholding step, which substantially mitigates this issue.

\section*{Acknowledgement}
YH and HZ are partially supported by an NSF IIS grant No. 2416897. HZ
would also like to thank the support from a Google Research Scholar Award. YH and HW are partially supported by a research grant from the Amazon-Illinois Center on AI for Interactive Conversational Experiences (AICE). The views and conclusions expressed in this paper are solely those of the authors and do not necessarily reflect the official policies or positions of the supporting companies and government agencies.

\bibliography{custom}

\begin{thebibliography}{58}
\providecommand{\natexlab}[1]{#1}

\bibitem[{Achiam et~al.(2023)Achiam, Adler, Agarwal, Ahmad, Akkaya, Aleman, Almeida, Altenschmidt, Altman, Anadkat et~al.}]{achiam2023gpt}
Josh Achiam, Steven Adler, Sandhini Agarwal, Lama Ahmad, Ilge Akkaya, Florencia~Leoni Aleman, Diogo Almeida, Janko Altenschmidt, Sam Altman, Shyamal Anadkat, et~al. 2023.
\newblock Gpt-4 technical report.
\newblock \emph{arXiv preprint arXiv:2303.08774}.

\bibitem[{Bai et~al.(2022{\natexlab{a}})Bai, Jones, Ndousse, Askell, Chen, DasSarma, Drain, Fort, Ganguli, Henighan et~al.}]{bai2022training}
Yuntao Bai, Andy Jones, Kamal Ndousse, Amanda Askell, Anna Chen, Nova DasSarma, Dawn Drain, Stanislav Fort, Deep Ganguli, Tom Henighan, et~al. 2022{\natexlab{a}}.
\newblock Training a helpful and harmless assistant with reinforcement learning from human feedback.
\newblock \emph{arXiv preprint arXiv:2204.05862}.

\bibitem[{Bai et~al.(2022{\natexlab{b}})Bai, Kadavath, Kundu, Askell, Kernion, Jones, Chen, Goldie, Mirhoseini, McKinnon et~al.}]{bai2022constitutional}
Yuntao Bai, Saurav Kadavath, Sandipan Kundu, Amanda Askell, Jackson Kernion, Andy Jones, Anna Chen, Anna Goldie, Azalia Mirhoseini, Cameron McKinnon, et~al. 2022{\natexlab{b}}.
\newblock Constitutional ai: Harmlessness from ai feedback.
\newblock \emph{arXiv preprint arXiv:2212.08073}.

\bibitem[{Bai et~al.(2024)Bai, Ying, Cao, Lv, He, Wang, Yu, Zeng, Xiao, Lyu et~al.}]{bai2024benchmarking}
Yushi Bai, Jiahao Ying, Yixin Cao, Xin Lv, Yuze He, Xiaozhi Wang, Jifan Yu, Kaisheng Zeng, Yijia Xiao, Haozhe Lyu, et~al. 2024.
\newblock Benchmarking foundation models with language-model-as-an-examiner.
\newblock \emph{Advances in Neural Information Processing Systems}, 36.

\bibitem[{Bradley and Terry(1952)}]{bradley1952rank}
Ralph~Allan Bradley and Milton~E Terry. 1952.
\newblock Rank analysis of incomplete block designs: I. the method of paired comparisons.
\newblock \emph{Biometrika}, 39(3/4):324--345.

\bibitem[{Cao et~al.(2021)Cao, Wong, and Lin}]{cao2021weak}
Zehong Cao, KaiChiu Wong, and Chin-Teng Lin. 2021.
\newblock Weak human preference supervision for deep reinforcement learning.
\newblock \emph{IEEE Transactions on Neural Networks and Learning Systems}, 32(12):5369--5378.

\bibitem[{Chapelle et~al.(2006)Chapelle, Sch{\"o}lkopf, and Zien}]{chapelle2006semi}
Olivier Chapelle, Bernhard Sch{\"o}lkopf, and Alexander Zien. 2006.
\newblock Semi-supervised learning. adaptive computation and machine learning series.

\bibitem[{Chen et~al.(2020)Chen, Wang, Tian, Yang, and Yang}]{chen-etal-2020-local}
Jiaao Chen, Zhenghui Wang, Ran Tian, Zichao Yang, and Diyi Yang. 2020.
\newblock \href {https://doi.org/10.18653/v1/2020.emnlp-main.95} {Local additivity based data augmentation for semi-supervised {NER}}.
\newblock In \emph{Proceedings of the 2020 Conference on Empirical Methods in Natural Language Processing (EMNLP)}, pages 1241--1251, Online. Association for Computational Linguistics.

\bibitem[{Chen et~al.(2021)Chen, Tworek, Jun, Yuan, Pinto, Kaplan, Edwards, Burda, Joseph, Brockman et~al.}]{chen2021evaluating}
Mark Chen, Jerry Tworek, Heewoo Jun, Qiming Yuan, Henrique Ponde de~Oliveira Pinto, Jared Kaplan, Harri Edwards, Yuri Burda, Nicholas Joseph, Greg Brockman, et~al. 2021.
\newblock Evaluating large language models trained on code.
\newblock \emph{arXiv preprint arXiv:2107.03374}.

\bibitem[{Christiano et~al.(2017)Christiano, Leike, Brown, Martic, Legg, and Amodei}]{christiano2017deep}
Paul~F Christiano, Jan Leike, Tom Brown, Miljan Martic, Shane Legg, and Dario Amodei. 2017.
\newblock Deep reinforcement learning from human preferences.
\newblock \emph{Advances in neural information processing systems}, 30.

\bibitem[{Cui et~al.(2023)Cui, Yuan, Ding, Yao, Zhu, Ni, Xie, Liu, and Sun}]{cui2023ultrafeedback}
Ganqu Cui, Lifan Yuan, Ning Ding, Guanming Yao, Wei Zhu, Yuan Ni, Guotong Xie, Zhiyuan Liu, and Maosong Sun. 2023.
\newblock Ultrafeedback: Boosting language models with high-quality feedback.
\newblock \emph{arXiv preprint arXiv:2310.01377}.

\bibitem[{Daniele and Suphavadeeprasit(2023)}]{daniele2023amplify-instruct}
Luigi Daniele and Suphavadeeprasit. 2023.
\newblock \href {https://huggingface.co/datasets/LDJnr/Capybara} {Amplify-instruct: Synthetically generated diverse multi-turn conversations for effecient llm training.}
\newblock \emph{arXiv preprint arXiv:(coming soon)}.

\bibitem[{Dong et~al.(2023)Dong, Xiong, Goyal, Zhang, Chow, Pan, Diao, Zhang, SHUM, and Zhang}]{dong2023raft}
Hanze Dong, Wei Xiong, Deepanshu Goyal, Yihan Zhang, Winnie Chow, Rui Pan, Shizhe Diao, Jipeng Zhang, KaShun SHUM, and Tong Zhang. 2023.
\newblock \href {https://openreview.net/forum?id=m7p5O7zblY} {{RAFT}: Reward ranked finetuning for generative foundation model alignment}.
\newblock \emph{Transactions on Machine Learning Research}.

\bibitem[{Dong et~al.(2024)Dong, Xiong, Pang, Wang, Zhao, Zhou, Jiang, Sahoo, Xiong, and Zhang}]{dong2024rlhf}
Hanze Dong, Wei Xiong, Bo~Pang, Haoxiang Wang, Han Zhao, Yingbo Zhou, Nan Jiang, Doyen Sahoo, Caiming Xiong, and Tong Zhang. 2024.
\newblock \href {https://arxiv.org/abs/2405.07863} {Rlhf workflow: From reward modeling to online rlhf}.
\newblock \emph{Preprint}, arXiv:2405.07863.

\bibitem[{Dubois et~al.(2024)Dubois, Li, Taori, Zhang, Gulrajani, Ba, Guestrin, Liang, and Hashimoto}]{dubois2024alpacafarm}
Yann Dubois, Chen~Xuechen Li, Rohan Taori, Tianyi Zhang, Ishaan Gulrajani, Jimmy Ba, Carlos Guestrin, Percy~S Liang, and Tatsunori~B Hashimoto. 2024.
\newblock Alpacafarm: A simulation framework for methods that learn from human feedback.
\newblock \emph{Advances in Neural Information Processing Systems}, 36.

\bibitem[{Ethayarajh et~al.(2022)Ethayarajh, Choi, and Swayamdipta}]{ethayarajh2022understanding}
Kawin Ethayarajh, Yejin Choi, and Swabha Swayamdipta. 2022.
\newblock Understanding dataset difficulty with $\mathcal{V}$-usable information.
\newblock In \emph{International Conference on Machine Learning}, pages 5988--6008. PMLR.

\bibitem[{Freund(1995)}]{freund1995boosting}
Yoav Freund. 1995.
\newblock Boosting a weak learning algorithm by majority.
\newblock \emph{Information and computation}, 121(2):256--285.

\bibitem[{Grandvalet and Bengio(2004)}]{grandvalet2004semi}
Yves Grandvalet and Yoshua Bengio. 2004.
\newblock Semi-supervised learning by entropy minimization.
\newblock \emph{Advances in neural information processing systems}, 17.

\bibitem[{Gulcehre et~al.(2023)Gulcehre, Paine, Srinivasan, Konyushkova, Weerts, Sharma, Siddhant, Ahern, Wang, Gu et~al.}]{gulcehre2023reinforced}
Caglar Gulcehre, Tom~Le Paine, Srivatsan Srinivasan, Ksenia Konyushkova, Lotte Weerts, Abhishek Sharma, Aditya Siddhant, Alex Ahern, Miaosen Wang, Chenjie Gu, et~al. 2023.
\newblock Reinforced self-training (rest) for language modeling.
\newblock \emph{arXiv preprint arXiv:2308.08998}.

\bibitem[{Gururangan et~al.(2019)Gururangan, Dang, Card, and Smith}]{gururangan-etal-2019-variational}
Suchin Gururangan, Tam Dang, Dallas Card, and Noah~A. Smith. 2019.
\newblock \href {https://doi.org/10.18653/v1/P19-1590} {Variational pretraining for semi-supervised text classification}.
\newblock In \emph{Proceedings of the 57th Annual Meeting of the Association for Computational Linguistics}, pages 5880--5894, Florence, Italy. Association for Computational Linguistics.

\bibitem[{He et~al.(2020)He, Gu, Shen, and Ranzato}]{He2020Revisiting}
Junxian He, Jiatao Gu, Jiajun Shen, and Marc'Aurelio Ranzato. 2020.
\newblock \href {https://openreview.net/forum?id=SJgdnAVKDH} {Revisiting self-training for neural sequence generation}.
\newblock In \emph{International Conference on Learning Representations}.

\bibitem[{He et~al.(2023)He, Wang, Li, and Zhao}]{he2023gradual}
Yifei He, Haoxiang Wang, Bo~Li, and Han Zhao. 2023.
\newblock \href {https://arxiv.org/abs/2310.13852} {Gradual domain adaptation: Theory and algorithms}.
\newblock \emph{Preprint}, arXiv:2310.13852.

\bibitem[{Huang et~al.(2024)Huang, Piqueres, Rasul, Schmid, Vila, and Tunstall}]{open_hermes_preferences}
Shengyi~Costa Huang, Agustín Piqueres, Kashif Rasul, Philipp Schmid, Daniel Vila, and Lewis Tunstall. 2024.
\newblock Open hermes preferences.
\newblock \url{https://huggingface.co/datasets/argilla/OpenHermesPreferences}.

\bibitem[{Ji et~al.(2024)Ji, Liu, Dai, Pan, Zhang, Bian, Chen, Sun, Wang, and Yang}]{ji2024beavertails}
Jiaming Ji, Mickel Liu, Josef Dai, Xuehai Pan, Chi Zhang, Ce~Bian, Boyuan Chen, Ruiyang Sun, Yizhou Wang, and Yaodong Yang. 2024.
\newblock Beavertails: Towards improved safety alignment of llm via a human-preference dataset.
\newblock \emph{Advances in Neural Information Processing Systems}, 36.

\bibitem[{Jiang et~al.(2023)Jiang, Ren, and Lin}]{jiang2023llm}
Dongfu Jiang, Xiang Ren, and Bill~Yuchen Lin. 2023.
\newblock Llm-blender: Ensembling large language models with pairwise ranking and generative fusion.
\newblock In \emph{Proceedings of the 61st Annual Meeting of the Association for Computational Linguistics (Volume 1: Long Papers)}, pages 14165--14178.

\bibitem[{Kumar et~al.(2020)Kumar, Ma, and Liang}]{kumar2020understanding}
Ananya Kumar, Tengyu Ma, and Percy Liang. 2020.
\newblock Understanding self-training for gradual domain adaptation.
\newblock In \emph{International conference on machine learning}, pages 5468--5479. PMLR.

\bibitem[{Lambert et~al.(2024)Lambert, Pyatkin, Morrison, Miranda, Lin, Chandu, Dziri, Kumar, Zick, Choi et~al.}]{lambert2024rewardbench}
Nathan Lambert, Valentina Pyatkin, Jacob Morrison, LJ~Miranda, Bill~Yuchen Lin, Khyathi Chandu, Nouha Dziri, Sachin Kumar, Tom Zick, Yejin Choi, et~al. 2024.
\newblock Rewardbench: Evaluating reward models for language modeling.
\newblock \emph{arXiv preprint arXiv:2403.13787}.

\bibitem[{Lee et~al.(2013)}]{lee2013pseudo}
Dong-Hyun Lee et~al. 2013.
\newblock Pseudo-label: The simple and efficient semi-supervised learning method for deep neural networks.
\newblock In \emph{Workshop on challenges in representation learning, ICML}, volume~3, page 896. Atlanta.

\bibitem[{Lee et~al.(2023)Lee, Phatale, Mansoor, Lu, Mesnard, Bishop, Carbune, and Rastogi}]{lee2023rlaif}
Harrison Lee, Samrat Phatale, Hassan Mansoor, Kellie Lu, Thomas Mesnard, Colton Bishop, Victor Carbune, and Abhinav Rastogi. 2023.
\newblock Rlaif: Scaling reinforcement learning from human feedback with ai feedback.
\newblock \emph{arXiv preprint arXiv:2309.00267}.

\bibitem[{Lewkowycz et~al.(2022)Lewkowycz, Andreassen, Dohan, Dyer, Michalewski, Ramasesh, Slone, Anil, Schlag, Gutman-Solo et~al.}]{lewkowycz2022solving}
Aitor Lewkowycz, Anders Andreassen, David Dohan, Ethan Dyer, Henryk Michalewski, Vinay Ramasesh, Ambrose Slone, Cem Anil, Imanol Schlag, Theo Gutman-Solo, et~al. 2022.
\newblock Solving quantitative reasoning problems with language models.
\newblock \emph{Advances in Neural Information Processing Systems}, 35:3843--3857.

\bibitem[{Li et~al.(2023)Li, Zhang, Dubois, Taori, Gulrajani, Guestrin, Liang, and Hashimoto}]{li2023alpacaeval}
Xuechen Li, Tianyi Zhang, Yann Dubois, Rohan Taori, Ishaan Gulrajani, Carlos Guestrin, Percy Liang, and Tatsunori~B Hashimoto. 2023.
\newblock Alpacaeval: An automatic evaluator of instruction-following models.

\bibitem[{Li et~al.(2022)Li, Choi, Chung, Kushman, Schrittwieser, Leblond, Eccles, Keeling, Gimeno, Dal~Lago et~al.}]{li2022competition}
Yujia Li, David Choi, Junyoung Chung, Nate Kushman, Julian Schrittwieser, R{\'e}mi Leblond, Tom Eccles, James Keeling, Felix Gimeno, Agustin Dal~Lago, et~al. 2022.
\newblock Competition-level code generation with alphacode.
\newblock \emph{Science}, 378(6624):1092--1097.

\bibitem[{Li et~al.(2019)Li, Peng, Zhang, Wang, and Si}]{li-etal-2019-semi-supervised-domain}
Zhenghua Li, Xue Peng, Min Zhang, Rui Wang, and Luo Si. 2019.
\newblock \href {https://doi.org/10.18653/v1/P19-1229} {Semi-supervised domain adaptation for dependency parsing}.
\newblock In \emph{Proceedings of the 57th Annual Meeting of the Association for Computational Linguistics}, pages 2386--2395, Florence, Italy. Association for Computational Linguistics.

\bibitem[{Lian et~al.(2023)Lian, Goodson, Pentland et~al.}]{lian2023openorca}
W~Lian, B~Goodson, E~Pentland, et~al. 2023.
\newblock Openorca: An open dataset of gpt augmented flan reasoning traces.

\bibitem[{Loshchilov and Hutter(2019)}]{loshchilov2018decoupled}
Ilya Loshchilov and Frank Hutter. 2019.
\newblock \href {https://openreview.net/forum?id=Bkg6RiCqY7} {Decoupled weight decay regularization}.
\newblock In \emph{International Conference on Learning Representations}.

\bibitem[{Meng et~al.(2020)Meng, Zhang, Huang, Xiong, Ji, Zhang, and Han}]{meng2020text}
Yu~Meng, Yunyi Zhang, Jiaxin Huang, Chenyan Xiong, Heng Ji, Chao Zhang, and Jiawei Han. 2020.
\newblock Text classification using label names only: A language model self-training approach.
\newblock In \emph{Proceedings of the 2020 Conference on Empirical Methods in Natural Language Processing (EMNLP)}, pages 9006--9017.

\bibitem[{Meta(2024)}]{llama3}
Meta. 2024.
\newblock \href {https://ai.meta.com/blog/meta-llama-3/} {Introducing meta llama 3: The most capable openly available llm to date.}

\bibitem[{Munos et~al.(2023)Munos, Valko, Calandriello, Azar, Rowland, Guo, Tang, Geist, Mesnard, Michi et~al.}]{munos2023nash}
R{\'e}mi Munos, Michal Valko, Daniele Calandriello, Mohammad~Gheshlaghi Azar, Mark Rowland, Zhaohan~Daniel Guo, Yunhao Tang, Matthieu Geist, Thomas Mesnard, Andrea Michi, et~al. 2023.
\newblock Nash learning from human feedback.
\newblock \emph{arXiv preprint arXiv:2312.00886}.

\bibitem[{Ouyang et~al.(2022)Ouyang, Wu, Jiang, Almeida, Wainwright, Mishkin, Zhang, Agarwal, Slama, Ray et~al.}]{ouyang2022training}
Long Ouyang, Jeffrey Wu, Xu~Jiang, Diogo Almeida, Carroll Wainwright, Pamela Mishkin, Chong Zhang, Sandhini Agarwal, Katarina Slama, Alex Ray, et~al. 2022.
\newblock Training language models to follow instructions with human feedback.
\newblock \emph{Advances in neural information processing systems}, 35:27730--27744.

\bibitem[{Park et~al.(2021)Park, Seo, Shin, Lee, Abbeel, and Lee}]{park2021surf}
Jongjin Park, Younggyo Seo, Jinwoo Shin, Honglak Lee, Pieter Abbeel, and Kimin Lee. 2021.
\newblock Surf: Semi-supervised reward learning with data augmentation for feedback-efficient preference-based reinforcement learning.
\newblock In \emph{International Conference on Learning Representations}.

\bibitem[{Rafailov et~al.(2024)Rafailov, Sharma, Mitchell, Manning, Ermon, and Finn}]{rafailov2024direct}
Rafael Rafailov, Archit Sharma, Eric Mitchell, Christopher~D Manning, Stefano Ermon, and Chelsea Finn. 2024.
\newblock Direct preference optimization: Your language model is secretly a reward model.
\newblock \emph{Advances in Neural Information Processing Systems}, 36.

\bibitem[{Schapire(1990)}]{schapire1990strength}
Robert~E Schapire. 1990.
\newblock The strength of weak learnability.
\newblock \emph{Machine learning}, 5:197--227.

\bibitem[{Schulman et~al.(2017)Schulman, Wolski, Dhariwal, Radford, and Klimov}]{schulman2017proximal}
John Schulman, Filip Wolski, Prafulla Dhariwal, Alec Radford, and Oleg Klimov. 2017.
\newblock Proximal policy optimization algorithms.
\newblock \emph{arXiv preprint arXiv:1707.06347}.

\bibitem[{Seeger(2000)}]{seeger2000learning}
Matthias Seeger. 2000.
\newblock Learning with labeled and unlabeled data.

\bibitem[{Sohn et~al.(2020)Sohn, Berthelot, Carlini, Zhang, Zhang, Raffel, Cubuk, Kurakin, and Li}]{sohn2020fixmatch}
Kihyuk Sohn, David Berthelot, Nicholas Carlini, Zizhao Zhang, Han Zhang, Colin~A Raffel, Ekin~Dogus Cubuk, Alexey Kurakin, and Chun-Liang Li. 2020.
\newblock Fixmatch: Simplifying semi-supervised learning with consistency and confidence.
\newblock \emph{Advances in neural information processing systems}, 33:596--608.

\bibitem[{Tarvainen and Valpola(2017)}]{tarvainen2017mean}
Antti Tarvainen and Harri Valpola. 2017.
\newblock Mean teachers are better role models: Weight-averaged consistency targets improve semi-supervised deep learning results.
\newblock \emph{Advances in neural information processing systems}, 30.

\bibitem[{Team et~al.(2024)Team, Mesnard, Hardin, Dadashi, Bhupatiraju, Pathak, Sifre, Rivi{\`e}re, Kale, Love et~al.}]{team2024gemma}
Gemma Team, Thomas Mesnard, Cassidy Hardin, Robert Dadashi, Surya Bhupatiraju, Shreya Pathak, Laurent Sifre, Morgane Rivi{\`e}re, Mihir~Sanjay Kale, Juliette Love, et~al. 2024.
\newblock Gemma: Open models based on gemini research and technology.
\newblock \emph{arXiv preprint arXiv:2403.08295}.

\bibitem[{Touvron et~al.(2023)Touvron, Martin, Stone, Albert, Almahairi, Babaei, Bashlykov, Batra, Bhargava, Bhosale et~al.}]{touvron2023llama}
Hugo Touvron, Louis Martin, Kevin Stone, Peter Albert, Amjad Almahairi, Yasmine Babaei, Nikolay Bashlykov, Soumya Batra, Prajjwal Bhargava, Shruti Bhosale, et~al. 2023.
\newblock Llama 2: Open foundation and fine-tuned chat models.
\newblock \emph{arXiv preprint arXiv:2307.09288}.

\bibitem[{Wang et~al.(2024)Wang, Dong, Delalleau, Zeng, Shen, Egert, Zhang, Sreedhar, and Kuchaiev}]{wang2024helpsteer2}
Zhilin Wang, Yi~Dong, Olivier Delalleau, Jiaqi Zeng, Gerald Shen, Daniel Egert, Jimmy~J. Zhang, Makesh~Narsimhan Sreedhar, and Oleksii Kuchaiev. 2024.
\newblock \href {https://arxiv.org/abs/2406.08673} {Helpsteer2: Open-source dataset for training top-performing reward models}.
\newblock \emph{Preprint}, arXiv:2406.08673.

\bibitem[{Wang et~al.(2023)Wang, Dong, Zeng, Adams, Sreedhar, Egert, Delalleau, Scowcroft, Kant, Swope et~al.}]{wang2023helpsteer}
Zhilin Wang, Yi~Dong, Jiaqi Zeng, Virginia Adams, Makesh~Narsimhan Sreedhar, Daniel Egert, Olivier Delalleau, Jane~Polak Scowcroft, Neel Kant, Aidan Swope, et~al. 2023.
\newblock Helpsteer: Multi-attribute helpfulness dataset for steerlm.
\newblock \emph{arXiv preprint arXiv:2311.09528}.

\bibitem[{Wei et~al.(2022)Wei, Wang, Schuurmans, Bosma, Xia, Chi, Le, Zhou et~al.}]{wei2022chain}
Jason Wei, Xuezhi Wang, Dale Schuurmans, Maarten Bosma, Fei Xia, Ed~Chi, Quoc~V Le, Denny Zhou, et~al. 2022.
\newblock Chain-of-thought prompting elicits reasoning in large language models.
\newblock \emph{Advances in neural information processing systems}, 35:24824--24837.

\bibitem[{Wu et~al.(2021)Wu, Ouyang, Ziegler, Stiennon, Lowe, Leike, and Christiano}]{wu2021recursively}
Jeff Wu, Long Ouyang, Daniel~M Ziegler, Nisan Stiennon, Ryan Lowe, Jan Leike, and Paul Christiano. 2021.
\newblock Recursively summarizing books with human feedback.
\newblock \emph{arXiv preprint arXiv:2109.10862}.

\bibitem[{Xie et~al.(2020)Xie, Luong, Hovy, and Le}]{xie2020self}
Qizhe Xie, Minh-Thang Luong, Eduard Hovy, and Quoc~V Le. 2020.
\newblock Self-training with noisy student improves imagenet classification.
\newblock In \emph{Proceedings of the IEEE/CVF conference on computer vision and pattern recognition}, pages 10687--10698.

\bibitem[{Yuan et~al.(2024{\natexlab{a}})Yuan, Cui, Wang, Ding, Wang, Deng, Shan, Chen, Xie, Lin et~al.}]{yuan2024advancing}
Lifan Yuan, Ganqu Cui, Hanbin Wang, Ning Ding, Xingyao Wang, Jia Deng, Boji Shan, Huimin Chen, Ruobing Xie, Yankai Lin, et~al. 2024{\natexlab{a}}.
\newblock Advancing llm reasoning generalists with preference trees.
\newblock \emph{arXiv preprint arXiv:2404.02078}.

\bibitem[{Yuan et~al.(2024{\natexlab{b}})Yuan, Pang, Cho, Sukhbaatar, Xu, and Weston}]{yuan2024self}
Weizhe Yuan, Richard~Yuanzhe Pang, Kyunghyun Cho, Sainbayar Sukhbaatar, Jing Xu, and Jason Weston. 2024{\natexlab{b}}.
\newblock Self-rewarding language models.
\newblock \emph{arXiv preprint arXiv:2401.10020}.

\bibitem[{Zhao et~al.(2023)Zhao, Joshi, Liu, Khalman, Saleh, and Liu}]{zhao2023slic}
Yao Zhao, Rishabh Joshi, Tianqi Liu, Misha Khalman, Mohammad Saleh, and Peter~J Liu. 2023.
\newblock Slic-hf: Sequence likelihood calibration with human feedback.
\newblock \emph{arXiv preprint arXiv:2305.10425}.

\bibitem[{Zheng et~al.(2024)Zheng, Chiang, Sheng, Zhuang, Wu, Zhuang, Lin, Li, Li, Xing et~al.}]{zheng2024judging}
Lianmin Zheng, Wei-Lin Chiang, Ying Sheng, Siyuan Zhuang, Zhanghao Wu, Yonghao Zhuang, Zi~Lin, Zhuohan Li, Dacheng Li, Eric Xing, et~al. 2024.
\newblock Judging llm-as-a-judge with mt-bench and chatbot arena.
\newblock \emph{Advances in Neural Information Processing Systems}, 36.

\bibitem[{Ziegler et~al.(2019)Ziegler, Stiennon, Wu, Brown, Radford, Amodei, Christiano, and Irving}]{ziegler2019fine}
Daniel~M Ziegler, Nisan Stiennon, Jeffrey Wu, Tom~B Brown, Alec Radford, Dario Amodei, Paul Christiano, and Geoffrey Irving. 2019.
\newblock Fine-tuning language models from human preferences.
\newblock \emph{arXiv preprint arXiv:1909.08593}.

\end{thebibliography}

\newpage

\appendix

\section{More experimental results} \label{appendix:exp}

\paragraph{Implementation}~We use the AdamW optimizer~\citep{loshchilov2018decoupled} with a learning rate of $5\times10^{-6}$ and a cosine learning rate schedule that includes 40 warmup steps. We use a context window of 3072 tokens with sample packing. The training process for each iteration is completed over one epoch, utilizing a global batch size of 128. In our implementation, each iteration starts from the same initial checkpoints (e.g., \gemmait for the Gemma experiments) instead of the checkpoint from the previous iteration, as training LLMs for more than one epoch is likely to lead to overfitting~\citep{wu2021recursively,ouyang2022training}, hurting the performance. This also ensures a fair comparison with the full SRM model reported at the end, as they both execute with one epoch. To ensure the reliability of our semi-supervised learning process, we choose a confidence threshold of 0.8. \looseness=-1

The experiments are run on NVIDIA A6000 GPUs with 48GB memory. In terms of running SRM on the full dataset, \pairrm requires 60 GPU hours, \gemma requires 20 GPU hours, and \llama requires 128 GPU hours.

\begin{table}[t!]
    \centering
    \scalebox{0.8}{
    \begin{tabular}{lc}
    \toprule
                  & \# Data (Pseudo-labeled portion)\\
    \midrule
    \llamait       & 0 (0)\\
    Partial SRM      & 43.75K (0)\\
    \midrule
    SSRM [t=1]      & 95.15K ($54.02\%$)\\
    SSRM [t=2]    & 290.75 ($84.95\%$)\\
    SSRM [t=3]    & 351.75 ($87.56\%$)\\
    \midrule
    Full SRM  & 700K (0)\\
    \bottomrule
    \end{tabular}}
    \caption{\llama models with number of data.}\label{tab:full_llama}
\end{table}

\begin{table}[t!]
    \centering
    \scalebox{0.8}{
    \begin{tabular}{lc}
    \toprule
                  & \# Data (Pseudo-labeled portion)\\
    \midrule
    \pairrm       & 0 (0)\\
    \midrule
    SSRM [t=1]      & 153K ($100\%$)\\
    SSRM [t=2]    & 198K ($100\%$)\\
    SSRM [t=3]    & 281K ($100\%$)\\
    \bottomrule
    \end{tabular}}
    \caption{\pairrm models with number of data.}\label{tab:full_pairrm}
\end{table}

\paragraph{Number of data used for training} In \Cref{tab:full_llama,tab:full_pairrm}, we report the number of data used in each iteration of SSRM for \llama and \pairrm. Similar to the findings as \Cref{tab:gemma}, with more iterations, an increasing number of pseudo-labeled data is included in the augmented dataset, as a result of the model's growing confidence in prediction. 

\begin{table}[t!]
    \centering
    \scalebox{0.65}{
    \begin{tabular}{lccccc}
    \toprule
                  & Chat   & Chat hard & Safety & Reasoning & Average \\
    \midrule
    \llamait        &  44.69 &  53.29    &  59.15 &  50.82    &  51.99  \\
    \midrule
    Partial SRM (1/16)        &  96.09 &  40.79    &  62.37 &  76.11    &  68.83  \\
    Partial SRM (1/4)        &  98.04 &  59.21    &  84.37 &  93.26    &  83.72  \\
    \midrule
    Full SRM &  98.60 &  65.35 &  88.81 &  92.07 &  86.21 \\
    \bottomrule
    \end{tabular}}
    \caption{Performance comparison for \llama using different number of data for SRM.}\label{tab:srm_data}
\end{table}

\paragraph{Number of data used for initial SRM} Here, we explain the reason to use different amount of data for the initial SRM for \gemma and \llama experiments. As shown in \Cref{tab:srm_data}, using only one-fourth of the data for SRM on \llamait already achieves an average performance only $2.5\%$ worse than that of the full SRM result, demonstrating saturation. In this case, we do not expect using more unlabeled data can continually enhance the performance, so we use one-sixteenth for the partial SRM instead, which leaves a larger room for improvement.

\section{Dataset Details}

HH-RLHF, UltraFeedback and UltraInteract are under MIT license. HelpSteer and PKU-SafeRLHF are under CC-BY-4.0 license. Distilabel-Capybara and Distilabel-Orca are under Apache-2.0 license. We cannot find license information for SHP and OpenHermesPreferences.

The data does not contain information that can be used to uniquely identifies individual people or offensive content.

\section{Potential Risks}
This paper presents work whose goal is to advance the field of NLP. There are many potential societal consequences of our work, none which we feel must be specifically highlighted here.


\end{document}